# Investigating the Effect of Emoji in Opinion Classification of Uzbek Movie Review Comments

Ilyos Rabbimov[1], Iosif Mporas[2], Vasiliki Simaki[3], Sami Kobilov[1]

[1] Applied Mathematics and Informatics Dept, Samarkand State University, Uzbekistan
[2] School of Engineering and Computer Science, University of Hertfordshire, United Kingdom
[3] Centre for Languages and Literature, Lund University, Sweden
`ilyos.rabbimov91@gmail.com, i.mporas@herts.ac.uk,`
`vasiliki.simaki@englund.lu.se, kobsam@yandex.ru`

**Abstract.** Opinion mining on social media posts has become more and more popular. Users often express their opinion on a topic not only with words but they also use image symbols such as emoticons and emoji. In this paper, we investigate the effect of emoji-based features in opinion classification of Uzbek texts, and more specifically movie review comments from YouTube. Several classification algorithms are tested, and feature ranking is performed to evaluate the discriminative ability of the emoji-based features.

## 1    Introduction

Over the last decades the use of Internet has dramatically increased with online activities like e-commerce, social media and blogs becoming extremely popular. Extraction of information from structured or unstructured online text is performed using text mining methodologies with applications among others in sentiment analysis, opinion mining [1, 2, 3], emotions [4] and stance classification [5]. In opinion classification, the positive or the negative opinion of users is automatically identified in data usually extracted from social media platforms such as Twitter, YouTube, Reddit and Facebook. The topics of discussion vary as well, and there are various studies that analyze movie reviews [6], political debates [7], the presence of offensive language in online texts [8] etc.  Most of the methodologies for automatic opinion classification that have been proposed in the literature are based on machine learning algorithms for classification, such as support vectors machines (SVMs) [1, 9, 10], Bayesian classifiers [1, 11], decision trees [1, 6] and neural networks [1, 12, 13]. In those approaches, the users' posts are represented by vectors of text features like language model based [1, 14, 15], word level [1, 6] and part-of-speech-based [1] statistical parameters. Other approaches are lexicon-based that rely on the presence in the data of words characterized as positive or negative [6, 12, 16, 17]. More recent approaches use word embeddings-based solutions to address the task [13, 14, 17].

In online text, and apart from the sentences consisting purely of word sequences, users frequently use emoticons or emoji to express themselves, emphasize and reinforce or mitigate the illocutionary force of their text. Emoticons are combinations of letters, numbers and symbols available on the keyboard, while emoji are pictures rather than typographic approximations of facial expressions, both expressing user's mood. The role of emoticons and emoji in sentiment analysis has been examined in previous studies. In [18], emoji sentiment ranking and sentiment map of the 751 most frequently used emoji was developed for automated sentiment analysis based on human annotators and tweets in 13 European languages. The authors found that most emoji are positive, especially the most popular ones, and the emotional perception of tweets changes significantly depending the presence or not of emoji in the text. In [19], a sociolinguistic study was presented exploiting emoji information for sentiment analysis. Similarly, in [20], the effect of emoji in sentiment analysis was studied, observing that taking into account emoji in sentiment analysis improves the recognition accuracy of the sentiment. The role of emoticons in the overall meaning of the text and their use in lexicon-based sentiment analysis was investigated in [21], where Dutch tweets and forum messages contained at least one emoticon were used. The results showed that the overall document polarity classification accuracy significantly improved when emoticons are considered. In [22], the problem of sarcasm detection in tweets using emoji was studied. In [23], the emoji2vec pre-trained embeddings of 1,661 unicode emoji was presented, and sentiment analysis on twitter posts was performed using word2vec and emoji2vec. In [24, 25], the effect of emoji in twitter posts was studied and in [26], the effect of using combined emoji-based features with textual features of Arabic tweets on sentiment classification task was presented.

While several corpora and language processing tools have been developed for the major languages of the world (i.e. languages spoken in many countries and/or by significant proportion of global population), not many language resources – that is the key component for any language-based technology – exist for other languages [27]. One of the less-resourced languages is Uzbek, which is the second most widely spoken Turkic language after the Turkish language. In Uzbek language until the first decades of the 20th century an Arabic-based script was used (the Yaña imlâ alphabet), then from 1928 to 1940 Latin-based Yañalif was used officially, and from 1940 the Cyrillic alphabet became the official script of the Uzbek language. In 1991, Uzbekistan's official script became the Yañalif-based Latin alphabet again. Despite the official status of the Latin script, the Cyrillic alphabet is still widespread and used in various occasions, especially by people who received education in the Soviet Union before the 1990's.

In the existing literature, there are natural language processing studies in the Uzbek language [28, 29, 30, 31, 32, 33, 34, 35], but very few can be found in Uzbek opinion mining. More specifically, in [36], the authors present an opinion mining dataset with 4,300 review comments about the top 100 applications from Google Play App Store used in Uzbekistan. The data is annotated according to the comment's positive/negative polarity, and baseline binary opinion classification results using support vector machines, logistic regression, recurrent neural networks and convolutional neural networks are presented. The same corpus and algorithms are used in [14], where the au-thors evaluate deep learning models for binary (positive vs. negative) opinion classification. In [37], a multilingual collection of sentiment lexicons including Uzbek is presented. In addition, there are few papers for sentiment/opinion mining in Turkish [38, 39, 40] and Kazakh [41, 42], which are languages belonging to the same language family. No emoji-related study in Uzbek opinion mining has been reported in the literature.

In this paper, we investigate the effect of emoji-based features in opinion classification of Uzbek movie review comments posted on YouTube. The reminder of the paper is organized as follows. In Section 2, the evaluated architecture for opinion classification is presented. In Section 3, the experimental setup is described. In Section 4, the experimental results are presented. Finally, Section 5 concludes our paper.

## 2 Uzbek Movie Reviews Opinion Classification

The architecture we used for the opinion classification of YouTube movie review comments and the investigation of the importance of emoji in the identification of user's opinion follows a standard approach. This approach is adopted in most opinion mining studies found in the bibliography and consists of the preprocessing of users' posts, the extraction of features and the classification experiments. The block diagram of the architecture is presented in Fig. 1.

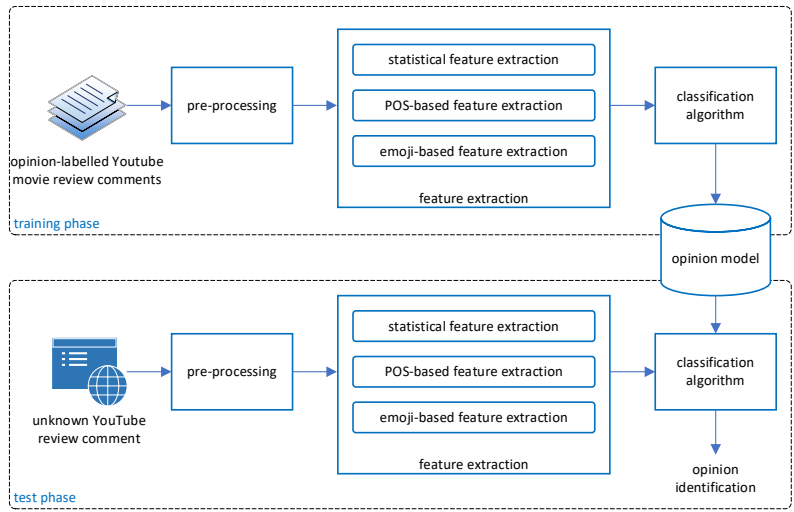

**Fig. 1.** Block diagram of the architecture for opinion classification of YouTube movie reviews, using text features based on word-based, part-of-speech-based and emoji-based statistics.

During the training phase, a set of movie review comments in Uzbek, extracted from YouTube, with known opinion labels are used to train an opinion classification model. The opinion labels are attributed manually by annotators that are native Uzbek speakers. More specifically, each review post is initially preprocessed and then features are extracted thus representing it by a feature vector. The text features are divided into statistical features, part-of-speech-based (henceforth, POS-based) features and emoji-based features, with the first ones consisting of statistical parameters on word level, the POS-based features consisting of statistical parameters on part-of-speech level and the last one consisting of statistical and opinion characteristic parameters of emoji. In the test phase, new movie reviews are preprocessed, and features are extracted similarly as in the training phase. The pretrained opinion classification model is used to classify the unknown movie review as positive or negative.

In the present evaluation, we have considered two opinion types, namely positive and negative, and not a scaled opinion ranking thus the classification models are binary. The architecture is generalized and thus can be used in multiclass opinion classification as well as in opinion mining of text from other Internet sources.

## 3 Experimental Setup

The architecture presented in Section 2 for opinion classification of Uzbek YouTube movie reviews was evaluated using the dataset, the text features and the classification algorithms described below.

### 3.1 Uzbek Dataset

The evaluation dataset is a collection of reviews of 75 Uzbek movies from YouTube using the YouTube Data API [43]. The posts are written in both Cyrillic and Latin scripts and the overall number of the review posts is 17,486 consisting of 121,941 words. For the annotation of the dataset, we employed six native Uzbek speakers. They annotated each review post into positive, negative or irrelevant, working individually, separately from each other. After completing the annotation of the whole set of posts given to them, and only for the posts for which the number of positive votes was equal to the number of negative ones, the annotators were asked to meet and jointly decide an opinion category. For the present evaluation, 2,044 positive and 519 negative posts were selected that contain at least one emoji. Statistical information about the dataset used in the present evaluation is presented in Table 1.

**Table 1.** Statistical information of the evaluation dataset.

| Characteristic | Statistical Information |
| --- | --- |
| # posts in Cyrillic | 1,453 |
| # posts in Latin | 1,100 |
| # posts in mixed Cyrillic – Latin | 10 |
| total # posts | 2,563 |
| # positive posts | 2,044 |
| # negative posts | 519 |
| minimum # emoji per post | 1 |
| maximum # emoji per post | 213 |
| average # emoji per post | 5.26 |

As mentioned in Section 1, the official script of Uzbekistan is the Latin-based alphabet, but the Cyrillic script is still widely used. For the purposes of our study, we manually converted all posts written in Cyrillic or mix with Latin into their corresponding Latin scripts. Also, due to lack of part-of-speech tagging tools for the Uzbek language, all words in the posts were manually tagged to their part-of-speech categories by two expert Uzbek linguists.

## 3.2 Feature Extraction

For each preprocessed movie review post, the statistical, POS-based and emoji-based features were calculated. More specifically, the calculated features are:

**Statistical Features.** Total number of characters; total number of characters without spaces; number of special characters ('(', ')', ' [', ']', '{', '}', '-', '/', '&', '|', '–', '—', '#', '%', '+', '*', '@', '$', '~', '=', '_', '«', '»', '<', '>', '^'); number of lower case characters; number of upper case characters; number of digits characters; number of all words; number of unique words; mean length of all unique words; maximum length of all words; minimum length of all words; mean length of all words; standard deviation of the length of all words; variance of the length of all words; kurtosis of the length of all words; skewness of the length of all words; percentile 25% of the length of all words; percentile 50% (median) of the length of all words; percentile 75% of the length of all words; number of punctuation characters ('.', ',', '!', '?', ':', ';'); number of words with length less than 4 characters; number of the hapax-legomena; number of the hapax-dislegomena.

**POS-based Features.** Number of nouns; number of proper nouns; number of verbs; number of adjectives; number of numerals; number of pronouns; number of adverbs; number of helping words; number of coordinating conjunctions; number of subordinating conjunctions of review; number of modal words; number of imitative words; number of interjections; number of auxiliaries; number of other words ($x$) like undefined or incomprehensible/meaningless cases.

**Emoji-based Features.** Number of emoji; average of sentiment score of all emoji per post [18]; number of positive emoji; number of negative emoji.

The number of statistical features is 23, the number of POS-based features is 15 and the number of emoji-based features is 4. The total dimensionality of the feature vector is equal to 42.

## 3.3 Classification Algorithms

For the classification stage, we used different well-known machine learning algorithms that are extensively used in several text classification tasks. In particular, we used:

**Instance based classifier (IBk).** A k-nearest neighbor classifier with linear search of the nearest neighbor and without weighting of the distance [44].

**Neural Networks (NN).** A multilayer perceptron neural network [45] with two hidden layers architecture (30 sigmoid nodes per hidden layer) trained with 5,000 iterations, using the back-propagation algorithm.

**Support Vector Machines (SVM).** The support vector machines using the sequential minimal optimization algorithm [46], which was tested using two different kernels, namely the radial basis kernel (rbf) and polynomial kernel (poly).

**Decision Trees.** Three tree algorithms were tested, namely the pruned C4.5 decision tree (J48) [47], the random forest (RandForest) [48] constructing a multitude of decision trees and the fast decision tree learner (RepTree) [49] that builds a decision tree using information gain or variance and prunes it using reduced-error pruning with back-fitting.

**Bayesian classifier (BayesNet).** We used the Bayes network learning [49], with simple estimator (alpha = 0.5) and the K2 search algorithm (maximum number of parents = 1), which is a probabilistic graphical model that represents a set of random variables and their conditional dependencies via a directed acyclic graph and the naive Bayes multinomial updateable, in which feature vectors represent the frequencies with which certain events have been generated by a multinomial.

All classifiers were implemented using the WEKA software [45]. For all algorithms, the free parameters were empirically selected, while parameter values not reported here were kept in their default values. For all classification algorithms two versions of opinion classification models were trained, one including emoji-based features (dimensionality equal to 42) and one without emoji-based features (dimensionality equal to 38).

## 4 Experimental Results

The architecture presented in Section 2 for opinion classification from YouTube movie review posts was evaluated according to the experimental setup described in Section 3. The performance of the evaluated algorithms was measured in terms of classification accuracy, i.e.

$$accuracy = \frac{TP+TN}{TP+FP+TN+FN}, \tag{1}$$

where *TP* are the true positives, *TN* are the true negatives, *FP* are the false positives and *FN* are the false negatives. The opinion classification accuracy for all evaluated classification algorithms and for two setups, namely with and without emoji-based text features, are presented in Table 2. In both setups, 10-fold cross validation was followed to avoid the overlap between the training and the test data. The best performance is indicated in bold.

As can be seen in Table 2, the best performing classification algorithm is the Random Forest (85.25%) when using emoji-based features, followed by REPTree and SVMs performing approximately 1% worse. The use of emoji-based features improved significantly the opinion classification accuracy across all evaluated algorithms. For the best performing classification algorithm (RandForest decision tree), the opinion classification accuracy improvement was approximately 5% when using the emoji-based features.

**Table 2.** Opinion classification accuracy for different classification algorithms, with and without emoji-based text features.

| Classification Algorithm | Accuracy (%) without emoji-based features | Accuracy (%) with emoji-based features |
|---|---|---|
| IBk | 75.89 | 80.26 |
| NN | 79.32 | 82.72 |
| SVM-poly | 79.63 | 84.55 |
| SVM-rbf | 79.75 | 84.39 |
| J48 | 77.10 | 83.46 |
| RandForest | 80.69 | **85.25** |
| REPTree | 78.62 | 84.12 |
| BayesNet | 64.30 | 75.34 |

In a further step, we evaluated the discriminative ability of the 42 calculated features using the ReliefF [50] feature ranking algorithm. The feature ranking results for the top-10 ranked features are shown in Table 3.

**Table 3.** Opinion classification feature ranking of the top 10 ranked features using the ReliefF algorithm.

| ranking | ReliefF score | feature name |
|---|---|---|
| 1 | 0.0671251 | average sentiment score of all emoji per post |
| 2 | 0.0192251 | skewness of the length of all words |
| 3 | 0.0185649 | number of adjectives |
| 4 | 0.017307 | minimum length of all words |
| 5 | 0.0162653 | maximum length of all words |
| 6 | 0.0157003 | mean length of all words |
| 7 | 0.0156075 | mean length of all unique words |
| 8 | 0.0148784 | percentile 25% of the length of all words |
| 9 | 0.014131 | number of words with length less than 4 characters |
| 10 | 0.0139635 | percentile 50% (median) of the length of all words |

As can be seen in Table 3, the most discriminative feature is the 'average sentiment score of all emoji per post', indicating the importance of emoji-based information in opinion classification. The remaining three emoji-based features where ranked in positions 24 ('number of negative emoji', ReliefF score: 0.0034666), 30 ('number of positive emoji', ReliefF score: 0.0019034) and 31 ('number of emoji', ReliefF score: 0.0017307), i.e. not in the first ten ranked features, however having positive ReliefF score which indicated that they also carry information with respect to users' opinion in the YouTube posts. The results from Tables 2 and 3 show that the positive effect of emoticons and emoji in opinion classification of other languages reported in previous studies [18-26] is also valid in the case of Uzbek opinion mining.

## 5      Conclusion

The identification of user's opinion in social media posts is of increasing interest in the research community as well as in commercial applications and services. Users often use emoticons and emoji apart from words in order to express their opinions. We investigated the effect of emoji-based features in opinion classification of Uzbek text from movie review comments from YouTube, and we evaluated their discriminative ability using the ReliefF feature ranking algorithm. The experimental results showed that Random Forest decision tree outperformed with accuracy equal to 85.25%, and the emoji-based features improved the opinion classification accuracy by approximately 5%. Also, the feature ranking evaluation showed that the most discriminative feature is the 'average sentiment score of all emoji per post'. The results in Uzbek opinion mining are in agreement with previous similar studies in other languages.

## 6      Acknowledgement

The authors are thankful to the six annotators and two linguists who contributed to the creation and processing of the dataset used in this work. This work was partially supported by the "El-Yurt Umidi" Foundation under the Cabinet of Ministers of the Republic of Uzbekistan (Internship Programme no. ST-2019-0080).

## References

1. Koumpouri, A., Mporas, I., Megalooikonomou, V.: Opinion Recognition on Movie Reviews by Combining Classifiers. Speech and Computer (SPECOM 2015), Volume 9319 of the series Lecture Notes in Computer Science, 309-316 (2015).
2. Poria, S., Cambria, E., Gelbukh, A.: Aspect extraction for opinion mining with a deep convolutional neural network. Knowledge-Based Systems 108, 42-49 (2016).
3. Sun, S., Luo, C., Chen, J.: A review of natural language processing techniques for opinion mining systems. Information Fusion 36, 10-25 (2017).
4. Dvoynikova, A., Verkholyak, O., Karpov, A.: Analytical review of methods for identifying emotions in text data. In: Proceedings of the III International Conference on Language Engineering and Applied Linguistics (PRLEAL-2019), pp. 8-21. CEUR-WS, (2020).
5. Simaki, V., Paradis, C., Kerren, A.: Stance classification in texts from blogs on the 2016 British referendum. In: Proc. of SPECOM 2017, pp. 700-709, Springer, (2017).
6. Koumpouri, A., Mporas, I., Megalooikonomou, V.: Evaluation of Four Approaches for "Sentiment Analysis on Movie Reviews": The Kaggle Competition. In: Proceedings of the 16th Int. Conf. on Eng/ring App. of Neural Networks (INNS), pp. 1-5. ACM, (2015).
7. Simaki, V., Paradis, C., Skeppstedt, M., Sahlgren, M., Kucher, K., Kerren, A.: Annotating speaker stance in discourse: the Brexit Blog Corpus. Corpus Linguistics and Linguistic Theory 1, ahead-of-print (2017).
8. Çöltekin, Ç.: A Corpus of Turkish Offensive Language on Social Media. In: Proc. of the 12th Language Resources and Evaluation Conference, pp. 6176-6186. ELRA, (2020).
9. Sunitha, P.B., Joseph, S., Akhil P.V.: A Study on the Performance of Supervised Algorithms for Classification in Sentiment Analysis. In: TENCON 2019, pp. 1351-1356. IEEE (2019).

10. Rinaldi, E., Musdholifah, A.: FVEC-SVM for opinion mining on Indonesian comments of youtube video. In: Proc. of the 2017 ICoDSE, pp. 1-5. IEEE, Indonesia (2017).
11. Saif, H., Fernández, M., He, Y., Alani, H.: On stopwords, filtering and data sparsity for sentiment analysis of twitter. In: Proceedings of the Ninth International Conference on Language Resources and Evaluation (LREC 2014), pp. 810-817. ELRA, Iceland (2014).
12. Cunha, A.A.L., Costa, M.C., Pacheco, M.A.C.: Sentiment Analysis of YouTube Video Comments Using Deep Neural Networks. In: Proceedings of the 18th International Conference on Artificial Intelligence and Soft Computing, pp. 561-570. Springer, Cham (2019).
13. Sido, J., Konopík, M.: Curriculum Learning in Sentiment Analysis. In: Proceedings of the 21st International Conference on Speech and Computer, pp. 444-450. Springer (2019).
14. Kuriyozov, E., Matlatipov, S., Alonso, M., Gómez-Rodríguez, C.: Deep Learning vs. Classic Models on a New Uzbek Sentiment Analysis Dataset. In: Proc. of the Human Language Technologies as a Challenge for Computer Science and Linguistics, pp. 258-262. (2019).
15. Jagdale, R.S., Shirsat, V.S., Deshmukh, S.N.: Sentiment analysis on product reviews using machine learning techniques. In: Cognitive Informatics and Soft Computing, pp. 639-647. Springer, Singapore (2019).
16. Esuli, A., Sebastiani, F.: Sentiwordnet: A publicly available lexical resource for opinion mining. In Proceedings of the 5th Conference on Language Resources and Evaluation (LREC06), pp. 417-422. European Language Resources Association (ELRA), (2006).
17. Rezaeinia, S., Rahmani, R., Ghodsi, A., Veisi, H.: Sentiment analysis based on improved pre-trained word embeddings. Expert Systems with Applications 117, 139-147 (2019).
18. Novak, P.K., Smailović, J., Sluban, B. and Mozetič, I.: Sentiment of emojis. PloS one 10(12), (2015).
19. Guibon, G., Ochs, M. and Bellot, P.: From emojis to sentiment analysis. (2016).
20. Shiha, M., Ayvaz, S.: The effects of emoji in sentiment analysis. IJCEE, 9(1), (2017).
21. Hogenboom, A., Bal, D., Frasincar, F., Bal, M., de Jong, F., Kaymak, U.: Exploiting emoticons in sentiment analysis. In: Proceedings of the 28th annual ACM symposium on applied computing, pp. 703-710. ACM, New York (2013).
22. Karthik, V., Nair, D., Anuradha, J.: Opinion Mining on Emojis using Deep Learning Techniques. Procedia computer science 132, 167-173 (2018).
23. Eisner, B., Rocktäschel, T., Augenstein, I., Bošnjak, M., Riedel, S.: emoji2vec: Learning emoji representations from their description. arXiv preprint arXiv:1609.08359 (2016).
24. Dandannavar, P.S., Mangalwede, S.R., Deshpande, S.B.: Emoticons and Their Effects on Sentiment Analysis of Twitter Data. In: EAI International Conference on Big Data Innovation for Sustainable Cognitive Computing, pp. 191-201. Springer, Cham (2020).
25. Wegrzyn-Wolska, K., Bougueroua, L., Yu, H., Zhong, J.: Explore the effects of emoticons on Twitter sentiment analysis. Comput. Sci. Inf. Technol. 2, 65 (2016).
26. Al-Azani, S., El-Alfy, E.S.M.: Combining emojis with Arabic textual features for sentiment classification. In: Proceedings of the 9th International Conference on Information and Communication Systems (ICICS), pp. 139-144. IEEE (2018).
27. Besacier, L., Barnard, E., Karpov, A., Schultz, T.: Automatic speech recognition for under-resourced languages: A survey. Speech Communication 56, 85-100 (2014).
28. Li, X., Tracey, J., Grimes, S., Strassel, S.: Uzbek-English and Turkish-English Morpheme Alignment Corpora. In: Proc. of the 10th LREC'16, pp. 2925–2930. ELRA, Portorož (2016).
29. Baisa V., Suchomel, V.: Large Corpora for Turkic Languages and Unsupervised Morphological Analysis. In: Proceedings of the 8th International Conference on Language Resources and Evaluation (LREC'12), pp. 28-32. ELRA, Turkey (2012).

30. Ismailov, A. Jalil, M.M.A., Abdullah Z., Rahim N.H.A.: A comparative study of stemming algorithms for use with the Uzbek language. In: Proceedings of the 3rd International Conference on Computer and Information Sciences (ICCOINS), pp. 7-12. IEEE (2016).
31. Xu, R., Yang, Y., Liu, H., Hsi, A.: Cross-lingual Text Classification via Model Translation with Limited Dictionaries. In: Proceedings of the 25th ACM International on Conference on Information and Knowledge Management (CIKM'16), pp. 95–104. ACM, (2016).
32. Abdurakhmonova, N.: Dependency Parsing based on Uzbek Corpus. In: Proceedings of the International Conference on Language Technologies for All (LT4All), (2019).
33. Chew, Y.C., Mikami, Y., Marasinghe, C.A., Nandasara, S.T.: Optimizing n-gram order of an N-gram based language identification algorithm for 63 written languages. The International Journal on Advances in ICT for Emerging Regions (ICTer) 2(2), (2009).
34. Uzbek text corpora page of Sketch Engine, https://www.sketchengine.eu/corpora-and-languages/uzbek-text-corpora, last accessed 2020/06/10.
35. Kuriyozov, E., Doval Y., Gómez-Rodríguez, C.: Cross-Lingual Word Embeddings for Turkic Languages. In: Proc. of the 12th LREC 2020, pp. 4047-4055. ELRA, (2020).
36. Kuriyozov, E., Matlatipov, S.: Building a New Sentiment Analysis Dataset for Uzbek Language and Creating Baseline Models. Multidisciplinary Digital Publishing Institute Proceedings 21(1), 37 (2019).
37. Chen Y., Skiena, S.: Building Sentiment Lexicons for All Major Languages. In: Proc. of the 52nd Annual Meeting of the Association for Computational Linguistics (Volume 1: Long Papers), pp. 383-389. ACL, Baltimore (2014).
38. Kaya, M., Guven, F., Toroslu, I.H.: Sentiment analysis of Turkish political news. In: Proceedings of the 2012 IEEE/WIC/ACM International Joint Conferences on Web Intelligence and Intelligent Agent Technology (Vol. 1), pp. 174-180. IEEE (2012).
39. Dehkharghani, R., Yanikoglu, B., Saygin, Y., Oflazer, K.: Sentiment analysis in Turkish at different granularity levels. Natural Language Engineering 23(4), 535–559 (2017).
40. Vural, A.G., Cambazoglu, B.B., Senkul, P., Tokgoz, Z.O.: A framework for sentiment analysis in Turkish: Application to polarity detection of movie reviews in Turkish. Computer and Information Sciences III, 437–445 (2012).
41. Yergesh, B., Bekmanova, G., Sharipbay, A., Yergesh, M.: Ontology-based sentiment analysis of Kazakh sentences. In: Proc. of the Int. Conf. on Computational Science and Its Applications, pp. 669-677. Springer, Cham (2017).
42. Sakenovich, N., Zharmagambetov, A.: On One Approach of Solving Sentiment Analysis Task for Kazakh and Russian Languages Using Deep Learning. In: International Conference on Computational Collective Intelligence, pp. 537-545. (2016).
43. YouTube Data API documentation page, https://developers.google.com/youtube/v3/docs/commentThreads, last accessed 2020/06/10.
44. Aha, D., Kibler, D.: Instance-based learning algorithms. Machine Learning 6, 37–66 (1991).
45. Eibe Frank, Mark, A. Hall, Ian, H.: The WEKA Workbench. Online Appendix for "Data Mining: Practical Machine Learning Tools and Techniques", Morgan Kaufmann, (2016).
46. Keerthi, S.S., Shevade, S.K., Bhattacharyya, C., Murthy, K.R.K.: Improvements to Platt's SMO algorithm for SVM classifier design. Neural Computation 13(3), 637–649 (2001).
47. Quinlan, R.: C4.5: Programs for machine learning. Morgan Kaufmann Publishers, (1993).
48. Breiman, L.: Random Forests. Machine Learning 45, 5-32 (2001).
49. Bouckaert, R.R.: Bayesian networks in Weka. Technical Report 14/2004. Computer Science Department. University of Waikato (2004)
50. Robnik-Šikonja, M., Kononenko, I.: An adaptation of Relief for attribute estimation in regression. In: Machine Learning: Proc. of ICML'97, Vol. 5, pp. 296-304 (1997).